\documentclass[11pt,a4paper]{article}
\usepackage[hyperref]{acl2019}
\usepackage{times}
\usepackage{latexsym}
\usepackage{url}

\usepackage{graphicx}  
\usepackage{enumitem}
\usepackage{supertabular,booktabs}
\usepackage{xcolor}
\usepackage{color, colortbl}
\usepackage[explicit]{titlesec}
\usepackage{pifont}

\definecolor{LGray}{gray}{0.9}
\definecolor{Gray}{gray}{0.8}
\definecolor{DGray}{gray}{0.7}

\titlespacing{\paragraph}{%
  0em}{%              left margin
  0\baselineskip}{% space before (vertical)
 0\baselineskip}%   
\aclfinalcopy % Uncomment this line for the final submission
\def\aclpaperid{2516} %  Enter the acl Paper ID here

\title{Automatically Identifying Complaints in Social Media}

\author{
    Daniel Preo\c{t}iuc-Pietro \\ Bloomberg LP \\ {\tt dpreotiucpie@bloomberg.net} \And
    Mihaela G\u{a}man \\ Politehnica University of Bucharest \\ {\tt mpgaman@gmail.com}
    \AND
    Nikolaos Aletras \\ University of Sheffield \\ {\tt n.aletras@sheffield.ac.uk}
    }

\date{}

\begin{document}
\maketitle
\begin{abstract}
Complaining is a basic speech act regularly used in human and computer mediated communication to express a negative mismatch between reality and expectations in a particular situation. Automatically identifying complaints in social media is of utmost importance for organizations or brands to improve the customer experience or in developing dialogue systems for handling and responding to complaints. In this paper, we introduce the first systematic analysis of complaints in computational linguistics. We collect a new annotated data set of written complaints expressed in English on Twitter.\footnote{Data and code is available here: \url{https://github.com/danielpreotiuc/complaints-social-media}} We present an extensive linguistic analysis of complaining as a speech act in social media and train strong feature-based and neural models of complaints across nine domains achieving a predictive performance of up to 79 F1 using distant supervision.

\end{abstract}

\section{Introduction}

\textit{Complaining is a basic speech act used to express a negative mismatch between reality and expectations towards a state of affairs, product, organization or event}~\cite{Olshtain87}. Understanding the expression of complaints in natural language and automatically identifying them is of utmost importance for: (a) linguists to obtain a better understanding of the context, intent and types of complaints on a large scale; (b) psychologists to identify human traits underpinning complaint behavior and expression; (c) organizations and advisers to improve the customer service by identifying and addressing client concerns and issues effectively in real time, especially on social media; (d) developing downstream natural language processing (NLP) applications, such as dialogue systems that aim to automatically identify complaints.

However, complaining has yet to be studied using computational approaches. The speech act of complaining, as previously defined in linguistics research~\citep{Olshtain87} and adopted in this study, has as its core the concept of violated or breached expectations i.e., the person posting the complaint had their favorable expectations breached by a party, usually the one to which the complaint is addressed.

Complaints have been previously analyzed by linguists~\citep{vasquez2011complaints} as distinctly different from expressing negative sentiment towards an entity. Key to the definition of complaints is the expression of the breach of expectations. Table~\ref{tbl:examples} shows examples of tweets highlighting the differences between complaints and sentiment. The first example expresses the writer's breach of expectations about an item that was expected to arrive, but does not express negative sentiment toward the entity, while the second shows mixed sentiment and expresses a complaint about a tax that was introduced. The third statement is an insult that implies negative sentiment, but there are not enough cues to indicate any breach of expectations; hence, this cannot be categorized as a complaint.

\begin{table}[t!]
\centering
\resizebox{\columnwidth}{!}{
\scriptsize
\begin{tabular}{|p{5cm}|c|c|}
\hline
\rowcolor{Gray} \textbf{Tweet} & \textbf{C} & \textbf{S} \\
\hline
\textsl{@FC\_Help hi, I ordered a necklace over a week ago and it still hasn't arrived (...)} & \ding{51} &  \\
\hline
\textsl{@BootsUK I love Boots! Shame you're introducing a man tax of 7\% in 2018 :(} & \ding{51} & \ding{51}  \\
\hline
\textsl{You suck} &  & \ding{51} \\
\hline
\end{tabular}
}
\caption{Examples of tweets annotated for complaint (C) and sentiment (S).}
\label{tbl:examples}
\end{table}

This paper presents the first extensive analysis of complaints in computational linguistics. Our contributions include:
\begin{enumerate}[noitemsep,topsep=0pt,leftmargin=1em]
    \item The first publicly available data set of complaints extracted from Twitter with expert annotations spanning nine domains (e.g.,\ software, transport);
    \item An extensive quantitative analysis of the syntactic, stylistic and semantic linguistic features distinctive of complaints;
    \item Predictive models using a broad range of features and machine learning models, which achieve high predictive performance for identifying complaints in tweets of up to 79 F1;
    %\item Using  tweets  containing  complaint  related  hashtags  as distant supervision and use domain adaptation to further boost predictive performance.
    %\item A distant-supervised approach to collect tweets containing complaint related hashtags together with domain adaptation to further boost predictive performance;
    \item A distant supervision approach to collect data combined with domain adaptation to boost predictive performance.
    %\item Experiments across a range of nine diverse domains of complaints (e.g.,\ software, transport).
\end{enumerate}

\section{Related Work}

Complaints have to date received significant attention in linguistics and marketing research. \citet{Olshtain87} provide one of the early definitions of a complaint as when a speaker expects a favorable event to occur or an unfavorable event to be prevented and these expectations are breached. Thus, the discrepancy between the expectations of the complainer and the reality is the key component of identifying complaints.

Complaining is considered to be a distinct speech act, as defined by speech act theory~\citep{austin1975things,searle1969speech} which is central to the field of pragmatics. Complaints are either addressed to the party responsible for enabling the breach of expectations (direct complaints) or indirectly mention the party (indirect complaints)~\citep{boxer1993social}. Complaints are widely considered to be among the face-threatening acts~\citep{brown1987politeness} -- acts that aim to damage the face or self-esteem of the person or entity the act is directed at. The concept of face~\citep{goffman1967interaction} represents the public image specific of each person or entity and has two aspects: positive (i.e.,\ the desire to be liked) and negative face (i.e.,\ the desire to not be imposed upon).
Complaints can intrinsically threaten both positive and negative face. Positive face of the responsible party is affected by having enabled the breach of expectations. Usually, when a direct complaint is made, the illocutionary function of the complaint is to request for a correction or reparation for these events. Thus, this aims to affect negative face by aiming to impose an action to be undertaken by the responsible party.
Complaints usually co-occur with other speech acts such as warnings, threats, suggestions or advice~\citep{Olshtain87,cohen1993production}.

Previous linguistics research has qualitatively examined the types of complaints elicited via discourse completion tests (DCT)~\citep{trosborg1995interlanguage} and in naturally occurring speech~\citep{laforest2002scenes}. Differences in complaint strategies and expression were studied across cultures~\citep{cohen1993production} and socio-demographic traits~\citep{boxer1993complaints}. In naturally occurring text, the discourse structure of complaints has been studied in letters to editors~\citep{hartford2004models,ranosa2004discourse}. In the area of linguistic studies on computer mediated communication,~\citet{vasquez2011complaints} performed an analysis of 100 negative reviews on TripAdvisor, which showed that complaints in this medium often co-occur with other speech acts including positive and negative remarks, frequently make explicit references to expectations not being met and directly demand a reparation or compensation. \citet{meinl2013electronic} studied complaints in eBay reviews by annotating 200 reviews in English and German with the speech act sequence that makes up each complaint e.g., warning, annoyance (the annotations are not available publicly or after contacting the authors). \citet{lexicon18naacl} analyze which financial complaints submitted to the Consumer Financial Protection Bureau will receive a timely response. Most recently, \citet{complaints19naacl} studied customer support dialogues and predicted if these complaints will be escalated with a government agency or made public on social media.

To the best of our knowledge, the only previous work that tackles a concept defined as a complaint with computational methods is by~\citet{zhou2016linguistic} which studies Yelp reviews. However, they define a complaint as a `sentence with negative connotation with supplemental information'. This definition is not aligned with previous research in linguistics (as presented above) and represents only a minor variation on sentiment analysis. They introduce a data set of complaints, unavailable at the time of this submission, and only perform a qualitative analysis, without building predictive models for identifying complaints.

\section{Data}

To date, there is no available data set with annotated complaints as previously defined in linguistics~\citep{Olshtain87}. Thus, we create a new data set of written utterances annotated with whether they express a complaint. We use Twitter as the data source because (1) it represents a platform with high levels of self-expression;  and (2) users directly interact with other users or corporate brand accounts. Tweets are openly available and represent a popular option for data selection in other related tasks such as predicting sentiment~\citep{rosenthal2017semeval}, affect~\citep{SemEval2018Task1}, emotion analysis~\citep{mohammad2015using}, sarcasm~\citep{gonzalez2011identifying,bamman2015contextualized}, stance~\citep{mohammad2016semeval}, text-image relationship~\citep{textimage2019acl} or irony~\citep{van2016monday,cervone2017irony,SemEval2018Task3}.

\subsection{Collection}

We choose to manually annotate tweets in order to provide a solid benchmark to foster future research on this task.

Complaints represent a minority of the total written posts on Twitter. We use a data sampling method that increases the hit rate of complaints, following previous work on labeling infrequent linguistic phenomena such as irony~\citep{SemEval2018Task1}. 
Numerous companies use Twitter to provide customer service and address user complaints. We select tweets directed to these accounts as candidates for complaint annotation.
We manually assembled a list of 93 customer service handles. Using the Twitter API,\footnote{\url{https://developer.twitter.com/}} we collected all the tweets that are available to download (the most recent 3,200). We then identified all the original tweets to which the customer support handle responded. We randomly sample an equal number of tweets addressed to each customer support handle for annotation. Using this method, we collected 1,971 tweets to which the customer support handles responded. 

Further, we have also manually grouped the customer support handles in several high-level domains based on their industry type and area of activity. We have done this to enable analyzing complaints by domain and assess transferability of classifiers across domains. 
In related work on sentiment analysis, reviews for products from four different domains were collected across domains in a similar fashion~\citep{blitzer2007biographies}. All customer support handles grouped by category are presented in Table~\ref{t:handles}.

\begin{table*}[h!]
\footnotesize
\centering
\resizebox{\textwidth}{!}{
\begin{tabular}{|c|c|c|c|c|c|c|c|c|}
\hline
\rowcolor{Gray}
\textbf{Food \& Beverage} & \textbf{Apparel} & \textbf{Retail} & \textbf{Cars} & \textbf{Services} & \textbf{Software \& Online Services} & \textbf{Transport} & \textbf{Electronics} & \textbf{Other} \\
\hline 
ABCustomerCare & NeimanMarcus & HarrodsService & HondaCustSvc & GEICO\_Service & YelpSupport & AirAsiaSupport & AskPlayStation & BlackandDecker \\
ArbysCares & FC\_Help & BN\_Care & VWCares & Safaricom\_Care & UbisoftSupport & SEPTA\_Social & XBoxSupport & WhirlpoolCare \\
KFC\_UKI\_Help & Zara\_Care & WalmartHelp & ChryslerCares & VirginMedia & SqSupportUK & FreaterAnglia & LenovoSupport & NYTCare \\
McDonalds  & NBaStoreSupport & BootsHelp & SubaruCustCare & ThreeUKSupport & AWSSupport  & RailMinIndia & AppleSupport & WashPostHelp \\
PizzaHut & HM\_CustServ & WholeFoods & AlfaRomeoCares & KenyaPower\_Care & SHO\_Help & VirginTrains & Moto\_Support & MACCosmetics \\
 & SupportAtTommy & BestBuySupport &  & GeorgiaPower & TeamTurboTax & Delta & OnePlus\_Support & HolidayInn \\
 & BurberyService & IKEAUSSupport &  & UPShelp & DropboxSupport & British\_Airways & SamsungSupport &  \\
 & Nordstrom & AmazonHelp &  & ComcastCares & AdobeCare & JetBlue & FitbitSupport &  \\
 & DSGsupport & AskEBay &  & AOLSupportHelp & Uber\_Support & United & BeatsSupport &  \\
 & TopmanAskUs &  &  & EE & NortonSupport & AmericanAir & NvidiaCC &  \\
 & SuperDry\_Care &  &  & VodafoneIN & MediumSupport & SouthwestAir & HPSupport &  \\
 &  ASOS\_HereToHelp &  &  & BTcare & TwitterSupport &  & NikeSupport &  \\
 &  &  &  & HMRCCustomers & Hulu\_Support &  &  &  \\ 
 &  &  &  & DirecTVService & MicrosoftHelps &  &  &  \\ 
 \hline
\end{tabular}
}
\caption{List of customer support handles by domain. The domain is chosen based on the most frequent product or service the account usually receives complaints about (e.g., NikeSupport receives most complaints about the Nike Fitness Bands).}
\label{t:handles}
\end{table*}

We add to our data set randomly sampled tweets to ensure that there is a more representative and diverse set of tweets for feature analysis and to ensure that the evaluation does not disproportionally contain complaints. %This is also useful for feature analysis and evaluation on a more diverse set of tweets. 
We thus additionally sampled 1,478 tweets consisting of two groups of 739 tweets: the first group contains random tweets addressed to any other Twitter handle (at-replies) to match the initial sample, while the second group contains tweets not addressed to a Twitter handle.

As preprocessing, we anonymize all usernames present in the tweet and URLs and replace them with placeholder tokens. To extract the unigrams used as features, we use DLATK, which handles social media content and markup such as emoticons or hashtags~\cite{DLATKemnlp2017}. Tweets were filtered for English using langid.py~\citep{lui2012langid} and retweets were excluded.

\subsection{Annotation}

We create a binary annotation task for identifying if a tweet contains a complaint or not. Tweets are short and usually express a single thought. Therefore, we consider the entire tweet as a complaint if it contains at least one complaint speech act. For annotation, we adopt as the guideline a complaint definition similar to that from previous linguistic research~\citep{Olshtain87,cohen1993production}:  \textsl{``A complaint presents a state of affairs which breaches the writer's favorable expectation''}. 

Each tweet was labeled by two independent annotators, authors of the paper, with significant experience in linguistic annotation. After an initial calibration run of 100 tweets (later discarded from the final data set), each annotator labeled all 1,971 tweets independently. The two annotators achieved a Cohen's Kappa $\kappa=0.731$, which is in the upper part of the \textit{substantial} agreement band~\citep{artstein2008inter}. Disagreements were discussed and resolved between the annotators. In total, 1,232 tweets (62.4\%) are complaints and 739 are not complaints (37.6\%). The statistics for each category is in Table~\ref{t:stats}.

\begin{table}[t!]
	\begin{center}       
    \small
    \resizebox{\columnwidth}{!}{
		\begin{tabular}{|l|c|c|}
		    \hline
            \rowcolor{DGray}
    \textbf{Category} & \textbf{Complaints} & \textbf{Not Complaints} \\
            \hline
    Food \& Beverage & 95 & 35  \\
    \rowcolor{LGray} Apparel & 141 & 117 \\
    Retail & 124 & 75 \\
    \rowcolor{LGray} Cars & 67 & 25 \\
    Services & 207 & 130 \\
    \rowcolor{LGray} Software \& Online Services & 189 & 103 \\
    Transport & 139 & 109 \\
	\rowcolor{LGray} Electronics & 174 & 112 \\
    Other & 96 & 33 \\    
\hline
    \rowcolor{LGray} Total & 1232 & 739 \\
    		\hline
		\end{tabular}
		}
		\caption{Number of tweets annotated as complaints across the nine domains.}
		\label{t:stats}
	\end{center}
\end{table}

\section{Features}

In our analysis and predictive experiments, we use the following groups of features: generic linguistic features proven to perform well in text classification tasks~\cite{occupation15acl,preoctiuc2017beyond,volkova2017identifying,preoctiuc2018race} (unigrams, LIWC, word clusters), methods for predicting sentiment or emotion which have an overlap with complaints and complaint specific features which capture linguistic aspects typical of complaints~\cite{meinl2013electronic,politeness}:

\paragraph{Unigrams.}
~We use the bag-of-words approach to represent each tweet as a TF-IDF weighted distribution over the vocabulary consisting of all words present in at least two tweets (2,641 words).

\paragraph{LIWC.}
~Traditional psychology studies use dictionary-based approaches to representing text. The most popular method is based on Linguistic Inquiry and Word Count (LIWC)~\citep{liwc} consisting of 73 manually constructed lists of words~\citep{liwc15} including parts-of-speech, topical or stylistic categories. Each tweet is thus represented as a distribution over these categories.

\paragraph{Word2Vec Clusters.}
~An alternative to LIWC for identifying semantic themes in a tweet is to use automatically generated word clusters. These clusters can be thought of as \textit{topics} i.e., groups of words that are semantically and/or syntactically similar. The clusters help reduce the feature space and provide good interpretability~\citep{impact14eacl,occupation15acl,Preoctiuc2015income,socec,Aletras2018}. %To create these groups of words, we use an automatic method that leverages word co-occurrence patterns in large corpora by making use of the distributional hypothesis: similar words tend to co-occur in similar contexts~\citep{harris54}. We use the method from ~\citep{occupation15acl} to compute topics using word2vec similarity~\citep{mikolov13naacl} and spectral clustering~\citep{Shi00,VonLuxburg07}.
We follow~\citet{occupation15acl} to compute clusters using spectral clustering~\citep{Shi00} applied to a word-word similarity matrix weighted with the cosine similarity of the corresponding word embedding vectors~\citep{mikolov13naacl}. The clusters help reduce the feature space and provide good interpretability.\footnote{We have tried other alternatives to building clusters: using NPMI \citep{Bouma2009}, GloVe \citep{glove} and LDA~\citep{lda}.}
For brevity and clarity, we present experiments using 200 clusters as in~\citep{occupation15acl}. We aggregated all the words in a tweet and represent each tweet as a distribution of the fraction of tokens belonging to each cluster.

\paragraph{Part-of-Speech Tags.}
~We analyze part-of-speech tag usage to quantify the syntactic patterns associated with complaints and to enhance the representation of unigrams. We part-of-speech tag all tweets using the Twitter model of the Stanford Tagger~\citep{derczynski2013twitter}. In prediction experiments we supplement each unigram feature with their POS tag (e.g., \textit{I\_PRP}, \textit{bought\_VBN}). For feature analysis, we represent each tweet as a bag-of-words distribution over part-of-speech unigrams and bigrams in order to uncover regular syntactic patterns specific of complaints.

\paragraph{Sentiment \& Emotion Models.}
~We use existing sentiment and emotion analysis models to study their relationship to complaint annotations and to measure their predictive power on our complaint data set. If the concepts of negative sentiment and complaint were to coincide, standard sentiment prediction models that have access to larger sets of training data should be very competitive on predicting complaints. We test the following models:

\begin{itemize}[noitemsep,topsep=0pt,leftmargin=1em]
\item \textbf{MPQA}: We use the MPQA sentiment lexicon~\cite{wiebe2005annotating} to assign a positive and negative score to each tweet based on the ratio of tokens in a tweet which appear in the positive and negative MPQA lists respectively. These scores are used as features.
\item \textbf{NRC}: We use the word lexicon derived using crowd-sourcing from~\cite{Mohammad10,Mohammad13} for assigning to each tweet the proportion of tokens that have positive, negative and neutral sentiment, as well as one of eight emotions that include the six basic emotions of Ekman~\cite{ekman} (anger, disgust, fear, joy, sadness and surprise) plus trust and anticipation. All scores are used as features in prediction in order to maximize their predictive power.
\item \textbf{Volkova \& Bachrach (V\&B)}: We quantify positive, negative and neutral sentiment as well as the six Ekman emotions for each message using the model made available in~\cite{volkova2016inferring} and use them as features in predicting complaints. The sentiment model is trained on a data set of 19,555 tweets that combine all previously annotated tweets across seven public data sets. 
\item \textbf{VADER}: We use the outcome of the rule-based sentiment analysis model which has shown very good predictive performance on predicting sentiment in tweets~\cite{gilbert2014vader}.
\item \textbf{Stanford}: We quantify sentiment using the Stanford sentiment prediction model as described in~\cite{socher2013recursive}.
\end{itemize}

\paragraph{Complaint Specific Features.}
~The features in this category are inspired by linguistic aspects specific to complaints~\cite{meinl2013electronic}:

$\bullet$ \textbf{Request.} The illocutionary function of complaints is often that of requesting for a correction or reparation for the event that caused the breach of expectations~\cite{Olshtain87}. We explicitly predict if an utterance is a request using the model introduced in~\citep{politeness}. 

$\bullet$ \textbf{Intensifiers.} In order to increase the face-threatening effect a complaint has on the complainee, intensifiers are usually used by the person expressing the complaint~\citep{meinl2013electronic}. We use features derived from: (1) capitalization patterns often used online as an equivalent to shouting (e.g., number/percentage of capitalized words, number/percentage of words starting with capitals, number/percentage of capitalized letters); and (2) repetitions of exclamation marks, question marks or letters within the same token.

$\bullet$ \textbf{Downgraders and Politeness Markers.} In contrast to intensifiers, downgrading modifiers are used to reduce the face-threat involved when voicing a complaint, usually as part of a strategy to obtain a reparation for the breach of expectation~\citep{meinl2013electronic}. Downgraders are coded by several dictionaries: play down (e.g., \textit{i wondered if}), understaters (e.g., \textit{one little}), disarmers (e.g., \textit{but}), downtoners (e.g., \textit{just}) and hedges (e.g., \textit{somewhat}). Politeness markers have a similar effect to downgraders and include apologies (e.g., \textit{sorry}), greetings at the start, direct questions, direct start (e.g., \textit{so}), indicative modals (e.g., \textit{can you}), subjunctive modals (e.g., \textit{could you}), politeness markers (e.g., \textit{please})~\citep{svarova2008politeness} and politeness maxims (e.g., \textit{i must say}). Finally, we directly predict the politeness score of the tweet using the model presented in ~\citep{politeness}.

$\bullet$  \textbf{Temporal References.} Temporal references are often used in complaints to stress how long a complainer has been waiting for a correction or reparation from the addressee or to provide context for their complaint (e.g., mentioning the date in which they have bought an item)~\citep{meinl2013electronic}. We identify time expressions in tweets using SynTime, which achieved state-of-the-art results across on several benchmark data sets~\citep{zhong2017time}. We represent temporal expressions both as days elapsed relative to the day of the post and in buckets of different granularities (one day, week, month, year).

$\bullet$ \textbf{Pronoun Types.} Pronouns are used in complaints to reveal the personal involvement or opinion of the complainer and intensify or reduce the face-threat of the complaint based on the person or type of the pronoun~\citep{claridge,meinl2013electronic}. We split pronouns using dictionaries into: first person, second person, third person, demonstrative (e.g., \textit{this}) and indefinite (e.g., \textit{everybody}).

\section{Linguistic Feature Analysis}

This section presents a quantitative analysis of the linguistic features distinctive of tweets containing complains in order to gain linguistic insight into this task and data. We perform analysis of all previously described feature sets using univariate Pearson correlation~\cite{schwartz2013personality}. We compute correlations independently for each feature between its distribution across messages (features are first normalized to sum up to unit for each message) and a variable encoding if the tweet was annotated as a complaint or not.

Top unigrams and part-of-speech features specific of complaints and non-complaints are presented in Table~\ref{t:unigrams}. The top features for the LIWC categories and Word2Vec topics are presented in Table~\ref{t:topics}. All correlations shown in these tables are statistically significant at $p<.01$, with Simes correction for multiple comparisons.

\begin{table}[t!]
\footnotesize
\centering
%\resizebox{\columnwidth}{!}{
\begin{tabular}{|>{\arraybackslash}m{0.25\columnwidth}|>{\arraybackslash}m{0.1\columnwidth}|>{\arraybackslash}m{0.25\columnwidth}|>{\arraybackslash}m{0.1\columnwidth}|}
\hline
\rowcolor{Gray} \multicolumn{2}{l}{\small{\textbf{Complaints}}} & \multicolumn{2}{l}{\small{\textbf{Not Complaints}}} \\
\rowcolor{DGray}
\small{\textbf{Feature}} & \small{\textbf{$r$}} &  \small{\textbf{Feature}} & \small{\textbf{$r$}} \\
\rowcolor{DGray} \multicolumn{4}{l}{\textbf{Unigrams}} \\
\hline 
not & .154 & $<$URL$>$ & .150 \\
\rowcolor{LGray} my & .131 & ! & .082 \\
working & .124 & he & .069 \\
\rowcolor{LGray} still & .123 & thank & .067 \\
on & .119 & , & .064 \\
\rowcolor{LGray} can't & .113 & love & .064 \\
service & .112 & lol & .061 \\
\rowcolor{LGray} customer & .109 & you & .060 \\
why & .108 & great & .058 \\
\rowcolor{LGray} website & .107 & win & .058 \\
no & .104 & ' & .058 \\
\rowcolor{LGray} ? & .098 & she & .054 \\
fix & .093 & : & .053 \\
\rowcolor{LGray} won't & .092 & that & .053 \\
been & .090 & more & .052 \\
\rowcolor{LGray} issue & .089 & it & .052 \\
days & .088 & would & .051 \\
\rowcolor{LGray} error & .087 & him & .047 \\
is & .084 & life & .046 \\
\rowcolor{LGray} charged & .083 & good & .046 \\
\hline
\rowcolor{DGray} \multicolumn{4}{l}{\textbf{POS (Unigrams and Bigrams)}} \\
\hline 
VBN & .141 & UH & .104 \\
\rowcolor{LGray} \$ & .118 & NNP & .098 \\
VBZ & .114 & PRP & .076 \\
\rowcolor{LGray} NN\_VBZ & .114 & HT & .076 \\
PRP\$ & .107 & PRP\_. & .076 \\
\rowcolor{LGray} PRP\$\_NN & .105 & PRP\_RB & .067 \\
VBG & .093 & NNP\_NNP & .062 \\
\rowcolor{LGray} CD & .092 & VBP\_PRP & .054 \\
WRB\_VBZ & .084 & JJ & .053 \\
\rowcolor{LGray} VBZ\_VBN & .084 & DT\_JJ & .051 \\
\hline
\end{tabular}
%}
\caption{Features associated with complaint and non-complaint tweets, sorted by Pearson correlation ($r$) computed between the normalized frequency of each feature and the complaint label across all tweets. All correlations are significant at $p<.01$, two-tailed t-test, Simes corrected.}
\label{t:unigrams}
\end{table}

\begin{table*}[t!]
\scriptsize
\centering
\resizebox{\textwidth}{!}{
\begin{tabular}{|>{\arraybackslash}m{0.17\columnwidth}|>{\arraybackslash}m{0.73\columnwidth}|>{\arraybackslash}m{0.1\columnwidth}||>{\arraybackslash}m{0.17\columnwidth}|>{\arraybackslash}m{0.73\columnwidth}|>{\arraybackslash}m{0.1\columnwidth}|}
\hline
\rowcolor{Gray} \multicolumn{3}{l}{\small{\textbf{Complaints}}} & \multicolumn{3}{l}{\small{\textbf{Not Complaints}}} \\
\rowcolor{DGray}
\small{\textbf{Label}} & \small{\textbf{Words}} & \small{\textbf{$r$}} & \small{\textbf{Label}} & \small{\textbf{Words}} & \small{\textbf{$r$}} \\
\rowcolor{DGray} \multicolumn{6}{l}{\textbf{LIWC Features}} \\
\hline 
NEGATE & not, no, can't, don't, never, nothing, doesn't, won't & .271 & POSEMO & thanks, love, thank, good, great, support, lol, win & .185 \\
\rowcolor{LGray} RELATIV & in, on, when, at, out, still, now, up, back, new  & .225 & AFFECT & thanks, love, thank, good, great, support, lol & .111 \\
FUNCTION & the, i, to, a, my, and, you, for, is, in & .204 & SHEHE & he, his, she, her, him, he's, himself & .105 \\
\rowcolor{LGray} TIME & when, still, now, back, new, never, after, then, waiting & .186 & MALE & he, his, man, him, sir, he's, son & .086 \\
DIFFER & not, but, if, or, can't, really, than, other, haven't & .169 & FEMALE & she, her, girl, mom, ma, lady, mother, female, mrs & .084 \\
\rowcolor{LGray} COGPROC & not, but, how, if, all, why, or, any, need & .132 & ASSENT & yes, ok, awesome, okay, yeah, cool, absolutely, agree & .080 \\
\hline
\rowcolor{DGray} \multicolumn{6}{l}{\textbf{Word2Vec Clusters}} \\
\hline 
Cust. Service & service, customer, contact, job, staff, assist, agent & .136 & Gratitude & thanks, thank, good, great, support, everyone, huge, proud & .089 \\
\rowcolor{LGray} Order & order, store, buy, free, delivery, available, package & .128 & Family & old, friend, family, mom, wife, husband, younger & .063 \\
Issues & delayed, closed, between, outage, delay, road, accident & .122 & Voting & favorite, part, stars, model, vote, models, represent & .060 \\
\rowcolor{LGray} Time Ref. & been, yet, haven't, long, happened, yesterday, took & .122 & Contests & Christmas, gift, receive, entered, giveaway, enter, cards & .058 \\
Tech Parts & battery, laptop, screen, warranty, desktop, printer & .100 &  Pets & dogs, cat, dog, pet, shepherd, fluffy, treats & .054 \\
\rowcolor{LGray} Access & use, using, error, password, access, automatically, reset & .098 & Christian & god, shall, heaven, spirit, lord, belongs, soul, believers & .053 \\
\hline
\end{tabular}
}
\caption{Group text features associated with tweets that are complaints and not complaints. Features are sorted by Pearson correlation ($r$) between their each feature's normalized frequency and the outcome. We restrict to only the top six categories for each feature type. All correlations are significant at $p<.01$, two-tailed t-test, Simes corrected. Within each cluster, words are sorted by frequency in our data set. Labels for Word2Vec clusters are assigned by the authors.}
\label{t:topics}
\end{table*}

\paragraph{Negations.}~Negations are uncovered through unigrams (\textit{not}, \textit{no}, \textit{won't}) and the top LIWC category (\textit{NEGATE}). Central to complaining is the concept of breached expectations. Hence the complainers use negations to express this discrepancy and to describe their experience with the product or service that caused this.

\paragraph{Issues.}~Several unigrams (\textit{error}, \textit{issue}, \textit{working}, \textit{fix}) and a cluster (\textit{Issues}) contain words referring to issues or errors. However, words regularly describing negative sentiment or emotions are not one of the most distinctive features for complaints. On the other hand, the presence of terms that show positive sentiment or emotions (\textit{good}, \textit{great}, \textit{win}, \textit{POSEMO}, \textit{AFFECT}, \textit{ASSENT}) are among the top most distinctive features for a tweet not being labeled as a complaint. In addition, other words and clusters expressing positive states such as gratitude (\textit{thank}, \textit{great}, \textit{love}) or laughter (\textit{lol}) are also distinctive for tweets that are not complaints. 

Linguistics research on complaints in longer documents identified that complaints are likely to co-occur with other speech acts, including with expressions of positive or negative emotions~\citep{vasquez2011complaints}. In our data set, perhaps due to the particular nature of Twitter communication and the character limit, complainers are much more likely to not express positive sentiment in a complaint and do not regularly post negative sentiment. Instead, they choose to focus more on describing the issue regarding the service or product in an attempt to have it resolved.

\paragraph{Pronouns.}~Across unigrams, part-of-speech patterns and word clusters, we see a distinctive pattern emerging around pronoun usage. Complaints use more possessive pronouns, indicating that the user is describing personal experiences. A distinctive part-of-speech pattern common in complaints is possessive pronouns followed by nouns (\textit{PRP\$\_NN}) which refer to items of services possessed by the complainer (e.g., \textit{my account}, \textit{my order}).
Complaints tend to not contain personal pronouns (\textit{he}, \textit{she}, \textit{it}, \textit{him}, \textit{you}, \textit{SHEHE}, \textit{MALE}, \textit{FEMALE}), as the focus on expressing the complaint is on the self and the party the complaint is addressed to and not other third parties.

\paragraph{Punctuation.}~Question marks are distinctive of complaints, as many complaints are formulated as questions to the responsible party (e.g., \textit{why is this not working?}, \textit{when will I get my response?}). Complaints are not usually accompanied by exclamation marks. Although exclamation marks are regularly used for emphasis in the context of complaints, most complainers in our data set prefer not to use them perhaps in an attempt to address them in a less confrontational manner.

% Parts-of-speech not characteristic of complaints include interjections, determiners and adjectives.

\paragraph{Temporal References.}~Mentions of time are specific of complaints (\textit{been}, \textit{still}, \textit{on}, \textit{days}, \textit{Temporal References} cluster). Their presence is usually needed to provide context for the event that caused the breach of expectations. Another role of temporal references is to express dissatisfaction towards non-responsiveness of the responsible party in addressing their previous requests. In addition, the presence of verbs in past participle (\textit{VBN}) is the most distinctive part-of-speech pattern of complaints. These are used to describe actions completed in the past (e.g., \textit{i've bought}, \textit{have come}) in order to provide context for the complaint.

\paragraph{Verbs.}~Several part-of-speech patterns distinctive of complaints involve present verbs in third person singular (\textit{VBZ}). In general, these verbs are used in complaints to reference an action that the author expects to happen, but his expectations are breached (e.g., \textit{nobody is answering}). Verbs in gerund or present participle are used as a complaint strategy to describe  things that just happened to a user (e.g., \textit{got an email saying my service will be terminated}).

\paragraph{Topics.}~General topics typical of complaint tweets include requiring assistance or customer support. Several groups of words are much more likely to appear in a complaint, although not used to express complaints per se: about orders or deliveries (in the retail domain), about access (in complaints to service providers) and about parts of tech products (in tech). This is natural, as people are more likely to deliberately tweet about an order or tech parts if they want to complain about them. This is similar to sentiment analysis, where not only emotionally valenced words are predictive of sentiment.

\section{Predicting Complaints}

In this section, we experiment with different approaches to build predictive models of complaints from text content alone. We first experiment with feature based approaches including Logistic Regression classification with Elastic Net regularization (LR) \citep{Zou2005elastic}.\footnote{We use the Scikit Learn implementation~\citep{sklearn}.} 
%classifiers have demonstrated good predictive accuracy in similar linguistic tasks~\citep{mohammad2013nrc,bamman2015contextualized}.
We train the classifiers with all individual feature types.

\paragraph{Neural Methods.}
~For reference, we experiment with two neural architectures. In both architectures, tweets are represented as sequences of one-hot word vectors which are first mapped into embeddings. 
A multi-layer perceptron (MLP) network~\citep{Hornik1989} feeds the embedded representation ($E=200$) of the tweet (mean embedding of its constituent words) into a dense hidden layer ($D=100$) followed by a ReLU activation function and dropout (0.2). The output layer is one dimensional dense layer with a sigmoid activation function. The second architecture, a Long-Short Term Memory (LSTM)~\citep{Hochreiter1997} network, processes sequentially the tweet by modeling one word (embedding) at each time step followed by the same output layer as in MLP. The size of the hidden state of the LSTM is $L=50$. We train the networks using the Adam optimizer~\citep{Kingma2014} (learning rate is set to $0.01$) by minimizing the binary cross-entropy.

\paragraph{Experimental Setup.}
~We conduct experiments using a nested stratified 10-fold cross-validation, where nine folds are used for training and one for testing (i.e., outer loop). In the inner loop, we choose the model parameters\footnote{We tune the regularization term, $\alpha$ and the mixing parameter of the LR model. For the neural networks, we tune the size of the embedding $E$, the dense hidden layer $D$, the LSTM cells $L$ and the learning rate of the optimizer.} using a 3-fold cross-validation on the tweets from the nine folds of training data (from the outer loop). Train/dev/test splits for each experiment are released together with the data for replicability. We report predictive performance of the models as the mean accuracy, F1 (macro-averaged) and ROC AUC over the 10 folds~\cite{dietterich1998approximate}.

\begin{table}[t!]
	\begin{center}       
    \scriptsize
    \resizebox{\columnwidth}{!}{
		\begin{tabular}{|l|c|c|c|}
		    \hline
            \rowcolor{Gray}
	\textbf{Model} & \textbf{Acc}  & \textbf{F1}  & \textbf{AUC} \\
            \hline
    \rowcolor{LGray}
    Most Frequent Class & 64.2 & 39.1 & 0.500 \\
    \hline \hline
    \rowcolor{LGray}
    Logistic Regression &  &  &  \\
    \enskip Sentiment -- MPQA  & 64.2 & 39.1 & 0.499 \\
    \enskip Sentiment -- NRC  & 63.9 & 42.2 & 0.599 \\
    \enskip Sentiment -- V\&B  & 68.9 & 60.0 & 0.696 \\
    \enskip Sentiment -- VADER & 66.0 & 54.2 & 0.654 \\
    \enskip Sentiment -- Stanford & 68.0 & 55.6 & 0.696 \\
    \hline
    \enskip Complaint Specific (all) & 65.7 & 55.2 & 0.634 \\
    \quad \textsl{Request} &  64.2 & 39.1 & 0.583 \\
    \quad \textsl{Intensifiers} & 64.5 & 47.3 & 0.639 \\
    \quad \textsl{Downgraders}  & 65.4 & 49.8 & 0.615 \\
    \quad \textsl{Temporal References} & 64.2 & 43.7 & 0.535 \\
    \quad \textsl{Pronoun Types} & 64.1 & 39.1  & 0.545 \\
    \hline
    \enskip POS Bigrams & 72.2 & 66.8 & 0.756 \\
    \enskip LIWC & 71.6 & 65.8 & 0.784 \\
    \enskip Word2Vec Clusters & 67.7 & 58.3 & 0.738 \\
    \enskip Bag-of-Words &  79.8 & 77.5 & 0.866 \\ 
    \enskip All Features & \textbf{80.5} & \textbf{78.0} & \textbf{0.873} \\ 
%        \hline
    %FastText & 64.4 & \mg{.} & 77.0 \\
    \hline \hline
    \rowcolor{LGray}
    Neural Networks & & &\\
    \enskip MLP & 78.3 & 76.2  & 0.845 \\
    %GRU & 79.3 & 68.5  & 85.7  \\
    %\rowcolor{LGray}
    \enskip LSTM & 80.2 & 77.0 & 0.864 \\            
    	\hline
%    Linear Ensemble & 79.8  & 74.2  & 87.2 \\
		\end{tabular}
	}
		\caption{Complaint prediction results using \textbf{logistic regression} (with different types of linguistic features), \textbf{neural network} approaches and the \textbf{most frequent class} baseline. Best results are in bold.}
		\label{t:results}
	\end{center}
\end{table}

\paragraph{Results.}
~Results are presented in Table~\ref{t:results}. Most sentiment analysis models show accuracy above chance in predicting complaints. The best results are obtained by the Volkova \& Bachrach model (Sentiment -- V\&B) which achieves 60 F1. However, models trained using linguistic features on the training data obtain significantly higher predictive accuracy. Complaint specific features are predictive of complaints, but to a smaller extent than sentiment, reaching an overall 55.2 F1. From this group of features, the most predictive groups are intensifiers and downgraders. Syntactic part-of-speech features alone obtain higher performance than any sentiment or complaint feature group, showing the syntactic patterns discussed in the previous section hold high predictive accuracy for the task. The topical features such as the LIWC dictionaries (which combine syntactic and semantic information) and Word2Vec topics perform in the same range as the part of speech tags. However, best predictive performance is obtained using bag-of-word features, reaching an F1 of up to 77.5 and AUC of 0.866. Further, combining all features boosts predictive accuracy to 78 F1 and 0.864 AUC. We notice that neural network approaches are comparable, but do not outperform the best performing feature-based model, likely in part due to the training data size.

\begin{table}[t!]
	\begin{center}       
    %\scriptsize
    \resizebox{\columnwidth}{!}{
		\begin{tabular}{|l|c|c|c|}
		    \hline
            \rowcolor{Gray}
	\textbf{Model} & \textbf{Acc}  & \textbf{F1}  & \textbf{AUC} \\
            \hline
    Most Frequent Class & 64.2 & 39.1 & 0.500 \\
    \hline
    LR-All Features -- Original Data & 80.5 & 78.0 & 0.873 \\ 
    \hline
    Dist. Supervision + Pooling & 77.2 & 75.7 & 0.853 \\
    Dist. Supervision + EasyAdapt & \textbf{81.2} & \textbf{79.0} & \textbf{0.885} \\       
    		\hline
		\end{tabular}
	}
		\caption{Complaint prediction results using the original data set and distantly supervised data. All models are based on logistic regression with bag-of-word and Part-of-Speech tag features.}
		\label{t:res_dist}
	\end{center}
\end{table}

\paragraph{Distant Supervision.}
~
We explore the idea of identifying extra complaint data using distant supervision to further boost predictive performance. Previous work has demonstrated improvements on related tasks relying on weak supervision e.g.,\ in the form of tweets with related hashtags~\citep{bamman2015contextualized,volkova2016inferring,cliche2017bb_twtr}. Following the same procedure, seven hashtags were identified with the help of the training data to likely correspond to complaints: \#appallingcustomercare, \#badbusiness, \#badcustomerserivice, \#badservice, \#lostbusiness, \#unhappycustomer, \#worstbrand. Tweets were collected to contain these hashtags from a combination of the 1\% Twitter archive between 2012-2018 and by filtering tweets with these hashtags in real-time from Twitter REST API for three months. We collected in total 18,218 tweets (excluding retweets and duplicates) equated to complaints. As negative complaint examples, the same amount of tweets were sampled randomly from the same time interval. All hashtags were removed and the data was preprocessed identically as the annotated data set.

We experiment with two techniques for combining distantly supervised data with our annotated data. First, the tweets obtained through distant supervision are simply added to the annotated training data in each fold (\textbf{Pooling}). Secondly, as important signal may be washed out if the features are joined across both domains, we experiment with domain adaptation using the popular EasyAdapt algorithm~\citep{daume2007frustratingly} (\textbf{EasyAdapt}). Experiments use logistic regression with bag-of-word features enhanced with part-of-speech tags, because these performed best in the previous experiment.

Results presented in Table~\ref{t:res_dist} show that the domain adaptation approach further boosts F1 by 1 point to 79 (t-test, p$<$0.5) and ROC AUC by 0.012. However, simply pooling the data actually hurts predictive performance leading to a drop of more than 2 points in F1.

\paragraph{Domain Experiments}
~We assess the performance of models trained using the best method and features by using in training: (1) using only in-domain data (\textbf{In-Domain)}; (2) adding out-of-domain data into the training set (\textbf{Pooling}); and (3) combining in- and out-of-domain data with EasyAdapt domain adaptation (\textbf{EasyAdapt}). The experimental setup is identical to the one described in the previous experiments. Table~\ref{t:domainadapt} shows the model performance in macro-averaged F1 using the best performing feature set.

\begin{table}[t!]
\footnotesize
\centering
\resizebox{\columnwidth}{!}{
\begin{tabular}{|l|c|c|c|}
\hline
\rowcolor{Gray}
\textbf{Domain} & \textbf{In-Domain} & \textbf{Pooling} & \textbf{EasyAdapt} \\
\hline 
Food \& Beverage & 63.9  & 60.9  & \textbf{83.1}  \\ 
\rowcolor{LGray} Apparel & \textbf{76.2} & 71.1  & 72.5  \\ 
Retail & 58.8  & \textbf{79.7}  & \textbf{79.7}  \\ 
\rowcolor{LGray} Cars &  41.5 & 77.8  & \textbf{80.9}  \\ 
Services &  65.2 & 75.9  & \textbf{76.7}  \\ 
\rowcolor{LGray} Software & 61.3  & 73.4  & \textbf{78.7}  \\ 
Transport & 56.4  & \textbf{73.4}  &  69.8 \\ 
\rowcolor{LGray} Electronics & 66.2  & 73.0  &  \textbf{76.2} \\ 
Other &  42.4 & \textbf{82.8} & \textbf{82.8}  \\ 
 \hline
\end{tabular}
}
\caption{Performance of models in Macro F1 on tweets from each domain.}
\label{t:domainadapt}
\end{table}

Results show that, in all but one case, adding out-of-domain data helps predictive performance. The apparel domain is qualitatively very different from the others as a large number of complaints are about returns or the company not stocking items, hence leading to different features being important for prediction. Domain adaptation is beneficial the majority of domains, lowering performance on a single domain compared to data pooling. This highlights the differences in expressing complaints across domains. Overall, predictive performance is high across all domains, with the exception of transport. 

\subsection*{Cross Domain Experiments}

Finally, Table~\ref{t:crossdomain} presents the results of models trained on tweets from one domain and tested on all tweets from other domains, with additional models trained on tweets from all domains except the one that the model is tested on.

We observe that predictive performance is relatively consistent across all domains with two exceptions (`Food \& Beverage' consistently shows lower performance, while `Other' achieves higher performance) when using all the data available from the other domains. 
\renewcommand{\arraystretch}{1.2}
\begin{table}[t!]
\footnotesize
\centering
\resizebox{\columnwidth}{!}{
\begin{tabular}{|c|c|c|c|c|c|c|c|c|c|}
\hline
\rowcolor{Gray}
\textbf{Test} & F\&B & A & R & Ca & Se & So & T & E & O \\
\rowcolor{Gray}
\textbf{Train} &  &  &  &  &  &  &  &  & \\ 
\hline 
Food \& Bev. & --&58.1&52.5&66.4&59.7&58.9&54.1&61.4&53.7 \\ 
\rowcolor{LGray} Apparel & 63.9 &--&74.4&65.1&70.8&71.2&68.5&76.9&85.6 \\ 
Retail & 58.8& 74.4&--&70.1&72.6&69.9&68.7&69.6&82.7 \\ 
\rowcolor{LGray} Cars & 68.7&61.1&65.1&--&58.8&67.&59.3&62.9&68.2 \\ 
Services &  65. &74.2&75.8&74.&--&68.8&74.2&77.9&77.9 \\ 
\rowcolor{LGray} Software & 62. &74.2&68.&67.9&72.8&--&72.8&72.1&80.6 \\ 
Transport & 59.3&71.7&72.4&67.&74.6&75.&--&72.6&81.7 \\ 
\rowcolor{LGray} Electronics & 61.6&75.2&71.&68.&75.&69.9&68.2&--&78.7 \\ 
Other & 56.1&71.3&72.4&70.2&73.5&67.2&68.5&71.&-- \\ 
\hline
All & 70.3 & 77.7 & 79.5 & 82.0 & 79.6 & 80.1 & 76.8 & 81.7 & 88.2 \\ 
 \hline
\end{tabular}
}
\caption{Performance of models trained with tweets from one domain and tested on other domains. All results are reported in ROC AUC. The \textbf{All} line displays results on training on all categories except the category in testing.}
\label{t:crossdomain}
\end{table}

\section{Conclusions \& Future Work}

We presented the first computational approach using methods from computational linguistics and machine learning to modeling complaints as defined in prior studies in linguistics and pragmatics~\citep{Olshtain87}. To this end, we introduced the first data set consisting of English Twitter posts annotated with complaints across nine domains. 
We analyzed the syntactic patterns and linguistic markers specific of complaints. Then, we built predictive models of complaints in tweets using a wide range of features reaching up to 79\% Macro F1 (0.885 AUC) and conducted experiments using distant supervision and domain adaptation to boost predictive performance. We studied performance of complaint prediction models on each individual domain and presented results with a domain adaptation approach which overall improves predictive accuracy. All data and code is available to the research community to foster further research on complaints.

A predictive model for identification of complaints is useful to companies that wish to automatically gather and analyze complaints about a particular event or product. This would allow them to improve efficiency in customer service or to more cheaply gauge popular opinion in a timely manner in order to identify common issues around a product launch or policy proposal. 

In the future, we plan to identify the target of the complaint in a similar way to aspect-based sentiment analysis~\citep{pontiki2016semeval}. We plan to use additional context and conversational structure to improve performance and identify the socio-demographic covariates of expressing and phrasing complaints. Another research direction is to study the role of complaints in personal conversation or in the political domain, e.g.,\ predicting political stance in elections \cite{Tsakalidis2018}.

\section*{Acknowledgments}
Nikolaos Aletras is supported by an Amazon AWS Cloud Credits for Research award. 

\bibliography{complaints19acl}

\begin{thebibliography}{69}
\expandafter\ifx\csname natexlab\endcsname\relax\def\natexlab#1{#1}\fi

\bibitem[{Aletras and Chamberlain(2018)}]{Aletras2018}
Nikolaos Aletras and Benjamin~Paul Chamberlain. 2018.
\newblock {Predicting Twitter User Socioeconomic Attributes with Network and
  Language Information}.
\newblock In \emph{Proceedings of the 29th on Hypertext and Social Media}, HT,
  pages 20--24.

\bibitem[{Artstein and Poesio(2008)}]{artstein2008inter}
Ron Artstein and Massimo Poesio. 2008.
\newblock {Inter-coder agreement for Computational Linguistics}.
\newblock \emph{Computational Linguistics}, 34(4):555--596.

\bibitem[{Austin(1975)}]{austin1975things}
John~Langshaw Austin. 1975.
\newblock \emph{How to do Things with Words}.
\newblock Oxford University Press.

\bibitem[{Bamman and Smith(2015)}]{bamman2015contextualized}
David Bamman and Noah~A Smith. 2015.
\newblock {Contextualized Sarcasm Detection on Twitter}.
\newblock In \emph{Proceedings of the 9th International Conference on Weblogs
  and Social Media}, ICWSM, pages 574--577.

\bibitem[{Blei et~al.(2003)Blei, Ng, and Jordan}]{lda}
David~M. Blei, Andrew~Y. Ng, and Michael~I. Jordan. 2003.
\newblock {Latent Dirichlet Allocation}.
\newblock \emph{Journal of Machine Learning Research}, 3:993--1022.

\bibitem[{Blitzer et~al.(2007)Blitzer, Dredze, and
  Pereira}]{blitzer2007biographies}
John Blitzer, Mark Dredze, and Fernando Pereira. 2007.
\newblock {Biographies, Bollywood, Boom-Boxes and Blenders: Domain Adaptation
  for Sentiment Classification}.
\newblock In \emph{Proceedings of the 45th Annual Meeting of the Association of
  Computational Linguistics}, ACL, pages 440--447.

\bibitem[{Bouma(2009)}]{Bouma2009}
Gerlof Bouma. 2009.
\newblock Normalized (pointwise) mutual information in collocation extraction.
\newblock \emph{Proceedings of GSCL}, pages 31--40.

\bibitem[{Boxer(1993{\natexlab{a}})}]{boxer1993complaints}
Diana Boxer. 1993{\natexlab{a}}.
\newblock {Complaints as Positive Strategies: What the Learner Needs to Know}.
\newblock \emph{Tesol Quarterly}, 27(2):277--299.

\bibitem[{Boxer(1993{\natexlab{b}})}]{boxer1993social}
Diana Boxer. 1993{\natexlab{b}}.
\newblock {Social Distance and Speech Behavior: The Case of Indirect
  Complaints}.
\newblock \emph{Journal of Pragmatics}, 19(2):103--125.

\bibitem[{Brown and Levinson(1987)}]{brown1987politeness}
Penelope Brown and Stephen~C Levinson. 1987.
\newblock \emph{{Politeness: Some Universals in Language Usage}}, volume~4.
\newblock Cambridge University Press.

\bibitem[{Cervone et~al.(2017)Cervone, Stepanov, Celli, and
  Riccardi}]{cervone2017irony}
Alessandra Cervone, Evgeny~A Stepanov, Fabio Celli, and Giuseppe Riccardi.
  2017.
\newblock Irony detection: from the twittersphere to the news space.
\newblock In \emph{CLiC-it 2017-Italian Conference on Computational
  Linguistics}, volume 2006.

\bibitem[{Claridge(2007)}]{claridge}
Claudia Claridge. 2007.
\newblock {Constructing a Corpus from the Web: Message Boards}.
\newblock \emph{Language and Computers}, 59(87).

\bibitem[{Cliche(2017)}]{cliche2017bb_twtr}
Mathieu Cliche. 2017.
\newblock {BB\_twtr at SemEval-2017 Task 4: Twitter Sentiment Analysis with
  CNNs and LSTMs}.
\newblock In \emph{Proceedings of International Workshop on Semantic Evaluation
  (SemEval-2017)}, *SEM, pages 573--580.

\bibitem[{Cohen and Olshtain(1993)}]{cohen1993production}
Andrew~D Cohen and Elite Olshtain. 1993.
\newblock {The Production of Speech Acts by EFL Learners}.
\newblock \emph{Tesol Quarterly}, 27(1):33--56.

\bibitem[{Danescu-Niculescu-Mizil et~al.(2013)Danescu-Niculescu-Mizil, Sudhof,
  Jurafsky, Leskovec, and Potts}]{politeness}
Cristian Danescu-Niculescu-Mizil, Moritz Sudhof, Dan Jurafsky, Jure Leskovec,
  and Christopher Potts. 2013.
\newblock {A Computational Approach to Politeness with Application to Social
  Factors}.
\newblock In \emph{Proceedings of the 51st Annual Meeting of the Association
  for Computational Linguistics}, ACL, pages 250--259.

\bibitem[{Daum{\'e}~III(2007)}]{daume2007frustratingly}
Hal Daum{\'e}~III. 2007.
\newblock {Frustratingly Easy Domain Adaptation}.
\newblock In \emph{Proceedings of the 45th Annual Meeting of the Association
  for Computational Linguistics}, ACL, pages 256--263.

\bibitem[{Derczynski et~al.(2013)Derczynski, Ritter, Clark, and
  Bontcheva}]{derczynski2013twitter}
Leon Derczynski, Alan Ritter, Sam Clark, and Kalina Bontcheva. 2013.
\newblock {Twitter Part-of-Speech Tagging for All: Overcoming Sparse and Noisy
  Data}.
\newblock In \emph{Proceedings of the International Conference Recent Advances
  in Natural Language Processing}, RANLP, pages 198--206.

\bibitem[{Dietterich(1998)}]{dietterich1998approximate}
Thomas~G Dietterich. 1998.
\newblock Approximate statistical tests for comparing supervised classification
  learning algorithms.
\newblock \emph{Neural computation}, 10(7):1895--1923.

\bibitem[{Ekman(1992)}]{ekman}
Paul Ekman. 1992.
\newblock {An Argument for Basic Emotions}.
\newblock \emph{Cognition \& Emotion}, 6(3-4):169--200.

\bibitem[{Gilbert and Hutto(2014)}]{gilbert2014vader}
CJ~Gilbert and Eric Hutto. 2014.
\newblock {VADER: A Parsimonious Rule-based Model for Sentiment Analysis of
  Social Media Text}.
\newblock In \emph{Proceedings of the 8th International Conference on Weblogs
  and Social Media}, ICWSM, pages 216--225.

\bibitem[{Goffman(1967)}]{goffman1967interaction}
Erving Goffman. 1967.
\newblock \emph{Interaction Ritual: Essays on Face-to-Face Interaction}.
\newblock Aldine.

\bibitem[{Gonz{\'a}lez-Ib{\'a}nez et~al.(2011)Gonz{\'a}lez-Ib{\'a}nez, Muresan,
  and Wacholder}]{gonzalez2011identifying}
Roberto Gonz{\'a}lez-Ib{\'a}nez, Smaranda Muresan, and Nina Wacholder. 2011.
\newblock {Identifying Sarcasm in Twitter: A Closer Look}.
\newblock In \emph{Proceedings of the 49th Annual Meeting of the Association
  for Computational Linguistics: Human Language Technologies: Short
  Papers-Volume 2}, ACL, pages 581--586.

\bibitem[{Hartford and Mahboob(2004)}]{hartford2004models}
Beverly Hartford and Ahmar Mahboob. 2004.
\newblock {Models of Discourse in the Letter of Complaint}.
\newblock \emph{World Englishes}, 23(4):585--600.

\bibitem[{Hochreiter and Schmidhuber(1997)}]{Hochreiter1997}
Sepp Hochreiter and J{\"u}rgen Schmidhuber. 1997.
\newblock Long short-term memory.
\newblock \emph{Neural computation}, 9(8):1735--1780.

\bibitem[{Hornik et~al.(1989)Hornik, Stinchcombe, and White}]{Hornik1989}
Kurt Hornik, Maxwell Stinchcombe, and Halbert White. 1989.
\newblock Multilayer feedforward networks are universal approximators.
\newblock \emph{Neural networks}, 2(5):359--366.

\bibitem[{Kingma and Ba(2014)}]{Kingma2014}
Diederik~P Kingma and Jimmy Ba. 2014.
\newblock Adam: A method for stochastic optimization.
\newblock \emph{arXiv preprint arXiv:1412.6980}.

\bibitem[{Laforest(2002)}]{laforest2002scenes}
Marty Laforest. 2002.
\newblock {Scenes of Family Life: Complaining in Everyday Conversation}.
\newblock \emph{Journal of Pragmatics}, 34(10-11):1595--1620.

\bibitem[{Lampos et~al.(2016)Lampos, Aletras, Geyti, Zou, and Cox}]{socec}
Vasileios Lampos, Nikolaos Aletras, Jens~K. Geyti, Bin Zou, and Ingemar~J. Cox.
  2016.
\newblock {Inferring the Socioeconomic Status of Social Media Users Based on
  Behaviour and Language}.
\newblock In \emph{Advances in Information Retrieval}, pages 689--695.

\bibitem[{Lampos et~al.(2014)Lampos, Aletras, Preo\c{t}iuc-Pietro, and
  Cohn}]{impact14eacl}
Vasileios Lampos, Nikolaos Aletras, Daniel Preo\c{t}iuc-Pietro, and Trevor
  Cohn. 2014.
\newblock {Predicting and Characterising User Impact on Twitter}.
\newblock In \emph{{Proceedings of the 14th Conference of the European Chapter
  of the Association for Computational Linguistics}}, EACL, pages 405--413.

\bibitem[{Lui and Baldwin(2012)}]{lui2012langid}
Marco Lui and Timothy Baldwin. 2012.
\newblock {langid.py: An off-the-shelf Language Identification Tool}.
\newblock In \emph{Proceedings of the ACL 2012 system demonstrations}, ACL,
  pages 25--30.

\bibitem[{Meinl(2013)}]{meinl2013electronic}
Marja~E Meinl. 2013.
\newblock \emph{{Electronic Complaints: An Empirical Study on British English
  and German Complaints on eBay}}, volume~18.
\newblock Frank \& Timme GmbH.

\bibitem[{Mikolov et~al.(2013)Mikolov, Yih, and Zweig}]{mikolov13naacl}
Tomas Mikolov, Wen-tau Yih, and Geoffrey Zweig. 2013.
\newblock {Linguistic Regularities in Continuous Space Word Representations}.
\newblock In \emph{{Proceedings of the 2013 Annual Conference of the North
  American Chapter of the Association for Computational Linguistics}}, NAACL,
  pages 746--751.

\bibitem[{Mikolov et~al.(2018)Mikolov, Yih, and Zweig}]{lexicon18naacl}
Tomas Mikolov, Wen-tau Yih, and Geoffrey Zweig. 2018.
\newblock {Deconfounded Lexicon Induction for Interpretable Social Science}.
\newblock In \emph{{Proceedings of the 2018 Annual Conference of the North
  American Chapter of the Association for Computational Linguistics}}, NAACL,
  pages 746--751.

\bibitem[{Mohammad et~al.(2016)Mohammad, Kiritchenko, Sobhani, Zhu, and
  Cherry}]{mohammad2016semeval}
Saif Mohammad, Svetlana Kiritchenko, Parinaz Sobhani, Xiaodan Zhu, and Colin
  Cherry. 2016.
\newblock {Semeval-2016 task 6: Detecting Stance in Tweets}.
\newblock In \emph{Proceedings of the 10th International Workshop on Semantic
  Evaluation (SemEval-2016)}, *SEM, pages 31--41.

\bibitem[{Mohammad et~al.(2018)Mohammad, Bravo-Marquez, Salameh, and
  Kiritchenko}]{SemEval2018Task1}
Saif~M. Mohammad, Felipe Bravo-Marquez, Mohammad Salameh, and Svetlana
  Kiritchenko. 2018.
\newblock Semeval-2018 {T}ask 1: {A}ffect in tweets.
\newblock In \emph{Proceedings of International Workshop on Semantic Evaluation
  (SemEval-2018)}, *SEM, pages 1--17.

\bibitem[{Mohammad and Kiritchenko(2015)}]{mohammad2015using}
Saif~M Mohammad and Svetlana Kiritchenko. 2015.
\newblock {Using Hashtags to Capture Fine Emotion Categories from Tweets}.
\newblock \emph{Computational Intelligence}, 31(2):301--326.

\bibitem[{Mohammad and Turney(2010)}]{Mohammad10}
Saif~M. Mohammad and Peter~D. Turney. 2010.
\newblock {Emotions Evoked by Common Words and Phrases: Using Mechanical Turk
  to Create an Emotion Lexicon}.
\newblock In \emph{Proceedings of the Workshop on Computational Approaches to
  Analysis and Generation of Emotion in Text}, NAACL, pages 26--34.

\bibitem[{Mohammad and Turney(2013)}]{Mohammad13}
Saif~M. Mohammad and Peter~D. Turney. 2013.
\newblock {Crowdsourcing a Word-Emotion Association Lexicon}.
\newblock \emph{Computational Intelligence}, 29(3):436--465.

\bibitem[{Olshtain and Weinbach(1987)}]{Olshtain87}
Elite Olshtain and Liora Weinbach. 1987.
\newblock {Complaints: A Study of Speech Act Behavior among Native and
  Non-native Speakers of Hebrew}.
\newblock \emph{Bertuccelli-Papi, M. (Eds.), The Pragmatic Perspective:
  Selected Papers from the 1985 International Pragmatics Conference}, pages
  195--208.

\bibitem[{Pedregosa et~al.(2011)Pedregosa, Varoquaux, Gramfort, Michel,
  Thirion, Grisel, Blondel, Prettenhofer, Weiss, Dubourg et~al.}]{sklearn}
Fabian Pedregosa, Ga{\"e}l Varoquaux, Alexandre Gramfort, Vincent Michel,
  Bertrand Thirion, Olivier Grisel, Mathieu Blondel, Peter Prettenhofer, Ron
  Weiss, Vincent Dubourg, et~al. 2011.
\newblock {Scikit-learn: Machine Learning in Python}.
\newblock \emph{JMLR}, 12.

\bibitem[{Pennebaker et~al.(2015)Pennebaker, Booth, Boyd, and Francis}]{liwc15}
James~W. Pennebaker, Roger~J. Booth, Ryan~L. Boyd, and Martha~E. Francis. 2015.
\newblock \emph{{Linguistic Inquiry and Word Count: LIWC2015}}.
\newblock Austin, TX: Pennebaker Conglomerates.

\bibitem[{Pennebaker et~al.(2001)Pennebaker, Francis, and Booth}]{liwc}
James~W. Pennebaker, Martha~E. Francis, and Roger~J. Booth. 2001.
\newblock \emph{{Linguistic Inquiry and Word Count}}.
\newblock Mahway: Lawrence Erlbaum Associates.

\bibitem[{Pennington et~al.(2014)Pennington, Socher, and Manning}]{glove}
Jeffrey Pennington, Richard Socher, and Christopher~D. Manning. 2014.
\newblock {GloVe: Global Vectors for Word Representation}.
\newblock In \emph{Proceedings of the 2014 Conference on Empirical Methods in
  Natural Language Processing}, EMNLP, pages 1532--1543.

\bibitem[{Pontiki et~al.(2016)Pontiki, Galanis, Papageorgiou, Androutsopoulos,
  Manandhar, Mohammad, Al-Ayyoub, Zhao, Qin, De~Clercq
  et~al.}]{pontiki2016semeval}
Maria Pontiki, Dimitris Galanis, Haris Papageorgiou, Ion Androutsopoulos,
  Suresh Manandhar, AL-Smadi Mohammad, Mahmoud Al-Ayyoub, Yanyan Zhao, Bing
  Qin, Orph{\'e}e De~Clercq, et~al. 2016.
\newblock {SemEval-2016 Task 5: Aspect based Sentiment Analysis}.
\newblock In \emph{Proceedings of the 10th International Workshop on Semantic
  Evaluation (SemEval-2016)}, pages 19--30.

\bibitem[{Preo\c{t}iuc-Pietro et~al.(2015)Preo\c{t}iuc-Pietro, Lampos, and
  Aletras}]{occupation15acl}
Daniel Preo\c{t}iuc-Pietro, Vasileios Lampos, and Nikolaos Aletras. 2015.
\newblock {An Analysis of the User Occupational Class through Twitter Content}.
\newblock In \emph{Proceedings of the 53rd Annual Meeting of the Association
  for Computational Linguistics and the 7th International Joint Conference on
  Natural Language Processing}, ACL, pages 1754--1764.

\bibitem[{Preo\c{t}iuc-Pietro and Ungar(2018)}]{preoctiuc2018race}
Daniel Preo\c{t}iuc-Pietro and Lyle Ungar. 2018.
\newblock {User-Level Race and Ethnicity Predictors from Twitter Text}.
\newblock In \emph{Proceedings of the 27th International Conference on
  Computational Linguistics}, COLING, pages 1534--1545.

\bibitem[{Preo{\c{t}}iuc-Pietro et~al.(2017)Preo{\c{t}}iuc-Pietro, Liu,
  Hopkins, and Ungar}]{preoctiuc2017beyond}
Daniel Preo{\c{t}}iuc-Pietro, Ye~Liu, Daniel Hopkins, and Lyle Ungar. 2017.
\newblock {Beyond Binary Labels: Political Ideology Prediction of Twitter
  Users}.
\newblock In \emph{Proceedings of the 55th Annual Meeting of the Association
  for Computational Linguistics}, ACL, pages 729--740.

\bibitem[{Preo{\c{t}}iuc-Pietro et~al.(2015)Preo{\c{t}}iuc-Pietro, Volkova,
  Lampos, Bachrach, and Aletras}]{Preoctiuc2015income}
Daniel Preo{\c{t}}iuc-Pietro, Svitlana Volkova, Vasileios Lampos, Yoram
  Bachrach, and Nikolaos Aletras. 2015.
\newblock Studying user income through language, behaviour and affect in social
  media.
\newblock \emph{PloS one}, 10(9):e0138717.

\bibitem[{Ranosa-Madrunio(2004)}]{ranosa2004discourse}
Marilu Ranosa-Madrunio. 2004.
\newblock {The Discourse Organization of Letters of Complaint to Editors in
  Philippine English and Singapore English}.
\newblock \emph{Philippine Journal of Linguistics}, 35(2):67--97.

\bibitem[{Rosenthal et~al.(2017)Rosenthal, Farra, and
  Nakov}]{rosenthal2017semeval}
Sara Rosenthal, Noura Farra, and Preslav Nakov. 2017.
\newblock {SemEval-2017 Task 4: Sentiment analysis in Twitter}.
\newblock In \emph{Proceedings of the 11th International Workshop on Semantic
  Evaluation (SemEval-2017)}, *SEM, pages 502--518.

\bibitem[{Schwartz et~al.(2013)Schwartz, Eichstaedt, Kern, Dziurzynski,
  Ramones, Agrawal, Shah, Kosinski, Stillwell, and
  Seligman}]{schwartz2013personality}
H~Andrew Schwartz, Johannes~C Eichstaedt, Margaret~L Kern, Lukasz Dziurzynski,
  Stephanie~M Ramones, Megha Agrawal, Achal Shah, Michal Kosinski, David
  Stillwell, and Martin~EP Seligman. 2013.
\newblock {Personality, Gender, and Age in the Language of Social Media: The
  Open-vocabulary Approach}.
\newblock \emph{PloS ONE}, 8(9).

\bibitem[{Schwartz et~al.(2017)Schwartz, Giorgi, Sap, Crutchley, Eichstaedt,
  and Ungar}]{DLATKemnlp2017}
H.~Andrew Schwartz, Salvatore Giorgi, Maarten Sap, Patrick Crutchley, Johannes
  Eichstaedt, and Lyle Ungar. 2017.
\newblock {DLATK: Differential Language Analysis ToolKit}.
\newblock In \emph{Proceedings of the 2017 Conference on Empirical Methods in
  Natural Language Processing: System Demonstrations}, EMNLP, pages 55--60.

\bibitem[{Searle(1969)}]{searle1969speech}
John~R Searle. 1969.
\newblock \emph{Speech Acts: An Essay in the Philosophy of Language}, volume
  626.
\newblock Cambridge University Press.

\bibitem[{Shi and Malik(2000)}]{Shi00}
Jianbo Shi and Jitendra Malik. 2000.
\newblock {Normalized Cuts and Image Segmentation}.
\newblock \emph{Transactions on Pattern Analysis and Machine Intelligence},
  22(8):888--905.

\bibitem[{Socher et~al.(2013)Socher, Perelygin, Wu, Chuang, Manning, Ng, and
  Potts}]{socher2013recursive}
Richard Socher, Alex Perelygin, Jean Wu, Jason Chuang, Christopher~D Manning,
  Andrew Ng, and Christopher Potts. 2013.
\newblock {Recursive Deep Models for Semantic Compositionality over a Sentiment
  Treebank}.
\newblock In \emph{Proceedings of the 2013 Conference on Empirical Methods in
  Natural Language Processing}, EMNLP, pages 1631--1642.

\bibitem[{Svarova(2008)}]{svarova2008politeness}
Jana Svarova. 2008.
\newblock \emph{{Politeness Markers in Spoken Language}}.
\newblock Ph.D. thesis, Masarykova Univerzita.

\bibitem[{Trosborg(1995)}]{trosborg1995interlanguage}
Anna Trosborg. 1995.
\newblock \emph{{Interlanguage Pragmatics: Requests, Complaints, and
  Apologies}}, volume~7.
\newblock Walter de Gruyter.

\bibitem[{Tsakalidis et~al.(2018)Tsakalidis, Aletras, Cristea, and
  Liakata}]{Tsakalidis2018}
Adam Tsakalidis, Nikolaos Aletras, Alexandra~I Cristea, and Maria Liakata.
  2018.
\newblock Nowcasting the stance of social media users in a sudden vote: {T}he
  case of the {G}reek {R}eferendum.
\newblock CIKM, pages 367--376.

\bibitem[{Van~Hee et~al.(2016)Van~Hee, Lefever, and Hoste}]{van2016monday}
Cynthia Van~Hee, Els Lefever, and V{\'e}ronique Hoste. 2016.
\newblock {Monday Mornings are my Fave:)\# not Exploring the Automatic
  Recognition of Irony in English tweets}.
\newblock In \emph{Proceedings of COLING 2016, the 26th International
  Conference on Computational Linguistics: Technical Papers}, pages 2730--2739.

\bibitem[{Van~Hee et~al.(2018)Van~Hee, Lefever, and Hoste}]{SemEval2018Task3}
Cynthia Van~Hee, Els Lefever, and Veronique Hoste. 2018.
\newblock Semeval-2018 {T}ask 3: {I}rony detection in {E}nglish {T}weets.
\newblock In \emph{Proceedings of International Workshop on Semantic Evaluation
  (SemEval-2018)}, *SEM, pages 39--50.

\bibitem[{V{\'a}squez(2011)}]{vasquez2011complaints}
Camilla V{\'a}squez. 2011.
\newblock {Complaints Online: The case of TripAdvisor}.
\newblock \emph{Journal of Pragmatics}, 43(6):1707--1717.

\bibitem[{Vempala and Preo{\c{t}}iuc-Pietro(2019)}]{textimage2019acl}
Alakananda Vempala and Daniel Preo{\c{t}}iuc-Pietro. 2019.
\newblock {Categorizing and Inferring the Relationship between the Text and
  Image of Twitter Posts}.
\newblock In \emph{Proceedings of the 57th Annual Meeting of the Association
  for Computational Linguistics}, ACL.

\bibitem[{Volkova and Bachrach(2016)}]{volkova2016inferring}
Svitlana Volkova and Yoram Bachrach. 2016.
\newblock {Inferring Perceived Demographics from User Emotional Tone and
  User-Environment Emotional Contrast}.
\newblock In \emph{Proceedings of the 54th Annual Meeting of the Association
  for Computational Linguistics}, ACL, pages 1567--–1578.

\bibitem[{Volkova and Bell(2017)}]{volkova2017identifying}
Svitlana Volkova and Eric Bell. 2017.
\newblock {Identifying Effective Signals to Predict Deleted and Suspended
  Accounts on Twitter across Languages}.
\newblock ICWSM, pages 290--298.

\bibitem[{Wiebe et~al.(2005)Wiebe, Wilson, and Cardie}]{wiebe2005annotating}
Janyce Wiebe, Theresa Wilson, and Claire Cardie. 2005.
\newblock {Annotating Expressions of Opinions and Emotions in Language}.
\newblock \emph{Language Resources and Evaluation}, 39(2-3):165--210.

\bibitem[{Yang et~al.(2019)Yang, Tan, Lu, Cui, Li, Chen, Xiong, Wang, Li, Pei,
  and Lin}]{complaints19naacl}
Wei Yang, Luchen Tan, Chunwei Lu, Anqi Cui, Han Li, Xi~Chen, Kun Xiong, Muzi
  Wang, Ming Li, Jian Pei, and Jimmy Lin. 2019.
\newblock {Detecting Customer Complaint Escalation with Recurrent Neural
  Networks and Manually-Engineered Features}.
\newblock In \emph{{Proceedings of the 2019 Annual Conference of the North
  American Chapter of the Association for Computational Linguistics (Industry
  Track)}}, NAACL, pages 56--63.

\bibitem[{Zhong et~al.(2017)Zhong, Sun, and Cambria}]{zhong2017time}
Xiaoshi Zhong, Aixin Sun, and Erik Cambria. 2017.
\newblock {Time expression analysis and recognition using syntactic token types
  and general heuristic rules}.
\newblock In \emph{Proceedings of the 55th Annual Meeting of the Association
  for Computational Linguistics}, ACL, pages 420--429.

\bibitem[{Zhou and Ganesan(2016)}]{zhou2016linguistic}
Guangyu Zhou and Kavita Ganesan. 2016.
\newblock {Linguistic Understanding of Complaints and Praises in User Reviews}.
\newblock In \emph{Proceedings of the 7th Workshop on Computational Approaches
  to Subjectivity, Sentiment and Social Media Analysis (WASSA)}, NAACL, pages
  109--114.

\bibitem[{Zou and Hastie(2005)}]{Zou2005elastic}
Hui Zou and Trevor Hastie. 2005.
\newblock {Regularization and Variable Selection via the Elastic Net}.
\newblock \emph{Journal of the Royal Statistical Society: Series B (Statistical
  Methodology)}, 67(2):301--320.

\end{thebibliography}
\bibliographystyle{acl_natbib}

\end{document}